\documentclass[
]{ceurart}

\usepackage{listings}
\usepackage{listings}
\usepackage{xcolor}

\definecolor{codegreen}{rgb}{0,0.6,0}
\definecolor{codegray}{rgb}{0.5,0.5,0.5}
\definecolor{codepurple}{rgb}{0.58,0,0.82}
\definecolor{backcolour}{rgb}{0.95,0.95,0.92}
\definecolor{teal}{HTML}{008081} %newly added

\lstdefinestyle{mystyle}{
    backgroundcolor=\color{backcolour},   
    commentstyle=\color{codegreen},
    keywordstyle=\color{magenta},
    numberstyle=\tiny\color{codegray},
    stringstyle=\color{codepurple},
    basicstyle=\ttfamily\footnotesize,
    breakatwhitespace=false,         
    breaklines=true,                 
    captionpos=b,                    
    keepspaces=true,                 
    numbers=left,                    
    numbersep=5pt,                  
    showspaces=false,                
    showstringspaces=false,
    showtabs=false,                  
    tabsize=2
}

\begin{document}

\copyrightyear{2023}
\copyrightclause{Copyright for this paper by its authors.
  Use permitted under Creative Commons License Attribution 4.0
  International (CC BY 4.0).}

\conference{Forum for Information Retrieval Evaluation, December 15-18, 2023, Panjim, India}

\title{Decoding Concerns: Multi-label Classification of Vaccine Sentiments in Social Media}
%\tnotemark[1]
%\tnotetext[1]{You can use this document as the template for preparing your publication. We recommend using the latest version of the ceurart style.}

\author[1]{Somsubhra De}[%
orcid=0000-0003-3839-2575,
email=somsubhra@outlook.in,
url=https://linkedin.com/in/somsubhrad,
]
\cormark[1]
\fnmark[1]
\address[1]{Indian Institute of Technology Madras,
  Tamil Nadu, India}
\address[2]{Indian Institute of Technology Kharagpur, West Bengal, India}

\author[2]{Shaurya Vats}[%
orcid=0009-0006-0283-2064,
email=shauryavats56@gmail.com,
url=https://linkedin.com/in/shaurya-vats-6a521a231,
]
\fnmark[1]

\cortext[1]{Corresponding author.}
\fntext[1]{These authors contributed equally.}

\begin{abstract}
In the realm of public health, vaccination stands as the cornerstone for mitigating disease risks and controlling their proliferation. The recent COVID-19 pandemic has highlighted how vaccines play a crucial role in keeping us safe. However the situation involves a mix of perspectives, with skepticism towards vaccines prevailing for various reasons such as political dynamics, apprehensions about side effects, and more. The paper addresses the challenge of comprehensively understanding and categorizing these diverse concerns expressed in the context of vaccination. Our focus is on developing a robust multi-label classifier capable of assigning specific concern labels to tweets based on the articulated apprehensions towards vaccines. To achieve this, we delve into the application of a diverse set of advanced natural language processing techniques and machine learning algorithms including transformer models like BERT, state of the art GPT 3.5, Classifier Chains \& traditional methods like SVM, Random Forest, Naive Bayes. We see that the cutting-edge large language model outperforms all other methods in this context.
\end{abstract}


\maketitle

\section{Introduction}
In the wake of the global COVID-19 pandemic, misinformation and vaccine hesitancy have emerged as significant societal challenges. The spread of anti-vaccine sentiment on social media platforms has amplified concerns. We address the complex task of multi-label classification of anti-vax tweets, aiming to identify and categorize the diverse spectrum of vaccine-related misinformation in the era of COVID-19. Through experimental evaluation, we demonstrate the efficacy of our proposed approach - our results not only underscore the prevalence of vaccine related concerns in social media discussions but also shed light on the nuanced nature of these concerns. The work provides valuable insights for policymakers, health organizations to communicate and intervene where required and emphasizes the significance of utilizing real-time social media discussions to inform evidence-based strategies that address public concerns and promote informed decision-making.
\newline
All the materials, including code, datasets, and related resources for our work, are accessible in \href{https://github.com/somsubhra04/AISoMe\_2023}{this repository}.
\section{Understanding the Task}
Given a tweet content, it can be classified into categories from the list of the following 12 total classes (each of the classes refer to the different concerns or reasons against the use of vaccines) that we have:
\begin{itemize}
\item \textbf{Unnecessary} - The tweet indicates that vaccines are unnecessary or alternate cures are better
\item \textbf{Mandatory} - The tweet is against mandatory vaccination
\item \textbf{Pharma} - The tweet indicates that the Big Pharmaceutical companies are just trying to earn money, or it is against such companies in general because of their history
\item \textbf{Conspiracy} - The tweet suggests some deeper conspiracy, and not just that the Big Pharma want to make money (e.g. vaccines are being used to track people, COVID is a hoax)
\item \textbf{Political} - The tweet expresses concerns that the governments or politicians are pushing their own agenda though the vaccines
\item \textbf{Country} - The tweet is against some vaccine because of the country where it was developed or manufactured
\item \textbf{Rushed} - The tweet expresses concerns that the vaccines have not been tested properly or the published data is not accurate
\item \textbf{Ingredients} - The tweet expresses concerns about the ingredients present in the vaccines (e.g. fetal cells, chemicals) or the technology used (e.g. mRNA vaccines can change the DNA)
\item \textbf{Side-effect} - The tweet expresses concerns about the side effects of the vaccines, including deaths caused
\item \textbf{Ineffective} - The tweet expresses concerns that the vaccines are not effective enough and are useless
\item \textbf{Religious} - The tweet is against vaccines because of religious reasons
\item \textbf{None} - No specific reason stated in the tweet or some reason other than the given ones
\end{itemize}
As this task involves multi-label classification, each tweet can be mapped to more than one labels, contingent on the stances expressed within the text. Below are a few tweet examples, each accompanied by associated labels and detailed descriptions to provide better context.
\begin{table}[h]
\centering
\caption{Examples of Tweet Classifications}
%\begin{tabular}{|c|c|c|}
\begin{tabular}{|p{7cm}|p{3cm}|p{4cm}|}
\hline
\textbf{Tweet Text} & \textbf{Labels} & \textbf{Tweet Description} \\
\hline
@jeffmcnamee @Amanda77197114 @alexanderchee BREAKING: FDA announces \textcolor{violet}{2 deaths} of Pfizer vaccine trial participants from ``\textcolor{violet}{serious adverse} events.â€. Fed Up \textcolor{brown}{Democrats} Say \textcolor{purple}{NO to Forced Vaccines} in NY &  \textcolor{violet}{{\tt side-effect}}\newline\textcolor{purple}{{\tt mandatory}}\newline\textcolor{brown}{{\tt political}}&\textit{The tweet reports FDA's announcement of two deaths during Pfizer vaccine trials due to `serious adverse events' and mentions that members of the Democratic party in New York are expressing their strong opposition to the idea of mandatory or compulsory vaccinations.}\\
\hline
My Take On The new vaccines? 1. I \textcolor{olive}{don't trust pharmaceutical companies} 2. Only 61\% of public will get vaccinated 3. Problems with people with \textcolor{violet}{allergies} 4. They're not free 5. \textcolor{magenta}{No data} on longevity 6. \textcolor{violet}{Side effects?} 7. Children under 16 were \textcolor{magenta}{not tested} in the Pfizer \textcolor{magenta}{trial} https://t.co/vZ4ZPkroc4 &  \textcolor{violet}{{\tt side-effect}}\newline\textcolor{olive}{{\tt pharma}}\newline\textcolor{magenta}{{\tt rushed}}&\textit{The tweet expresses skepticism about new vaccines, citing distrust of pharmaceutical companies, concerns about side effects of rushed vaccines, and other issues, particularly focusing on the lack of comprehensive testing.}\\
\hline
Why would we get a vaccine with so many \textcolor{violet}{side effects}? Ccp covid has a \textcolor{teal}{99.7\% survival rate}, so \textcolor{teal}{why get} a vaccine? Thatâ€™s stupid and dangerous. &  \textcolor{violet}{{\tt side-effect}}\newline\textcolor{teal}{{\tt unnecessary}}&\textit{The tweet questions the need for vaccines due to perceived side effects and argues against it, citing a high survival rate.}\\
\hline
@TorontoStar Then I suppose there is no hope for a vaccine.   Nothing coming and it \textcolor{orange}{won't work} if it does come. It is Lockdowns Forever, or take your chances with Covid. Lockdowns are no life at all. My choice it to take my chances. Bring it on! &  \textcolor{orange}{{\tt ineffective}}&\textit{The tweet expresses skepticism about the effectiveness of vaccines, suggesting that there's no hope for them and indicating a preference for taking their chances with COVID-19 over enduring continuous lockdowns.}\\
\hline
Oh my! One doesn't have to be an expert at reading body language to know he's \textcolor{blue}{covering something} up. \textcolor{blue}{Depopulation} is his \textcolor{blue}{game}. &  \textcolor{blue}{{\tt conspiracy}}&\textit{The tweet implies a conspiracy claiming that someone is involved in an agenda, planning to reduce the global population intentionally.}\\
\hline
\end{tabular}
\end{table}
\section{Dataset}
The training dataset, comprising \textbf{9,921 anti-vaccine} tweet texts (posted during 2020-21) with corresponding tweet IDs and annotated labels, was sourced from the CAVES\textsuperscript{\cite{caves2022}} Dataset. The test set consists of 486 tweets (along with their IDs), encompassing various vaccine types such as MMR, flu vaccines and more, in addition to COVID-19 vaccines.
\subsection{Some insights about Train set}
Performing an EDA, we observe that the dataset exhibits class imbalance with certain labels having significantly higher counts compared to others. The two primary concerns {\tt side-effect} and {\tt ineffective} appear to dominate the dataset where as vaccine skepticism due to religious reasons and concerns related to the country of origin are less prevalent.
\begin{figure}
  \centering
  \includegraphics[width=\linewidth]{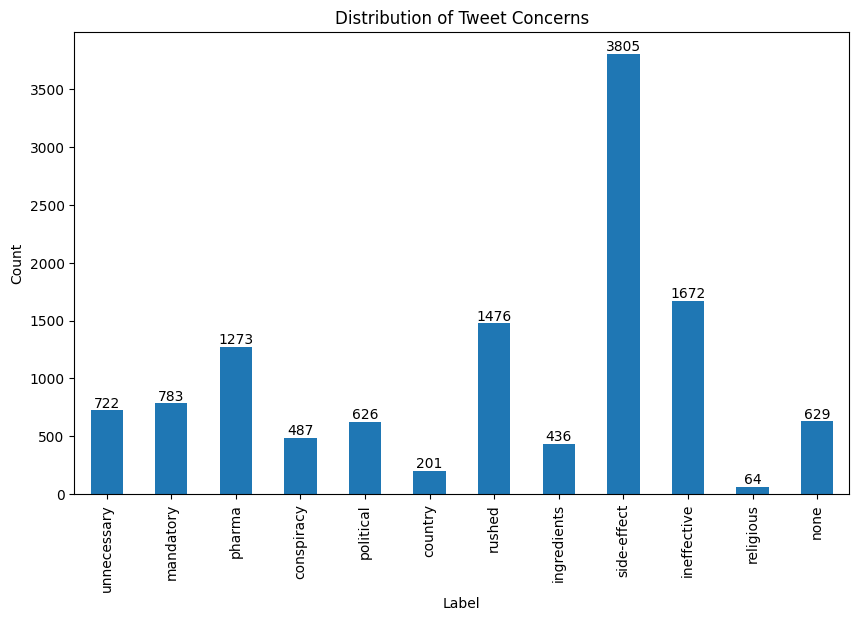}
  \caption{Categorical Data split (Train set)}
\end{figure}

\begin{figure}
  \centering
  \includegraphics[width=\linewidth]{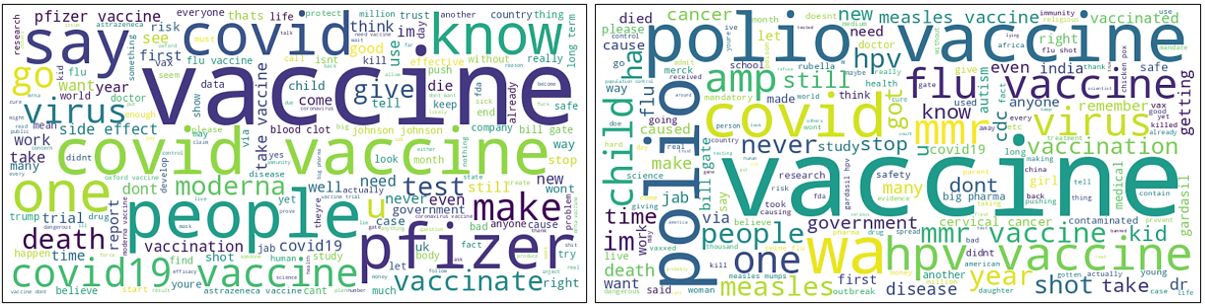}
  \caption{Word Cloud visualization of Train (left) vs. Test set Data (right)}
\end{figure}

\section{Our Methodology}
\subsection{Data Pre-processing}
To enhance the model's performance, we conducted comprehensive data pre-processing on both the training and test datasets. This involved meticulous cleaning to eliminate noisy data, and various other steps ensuring that the model's integrity remained intact.
\begin{itemize}
\item Stop word Removal: Elimination of commonly occurring English stop words (eg, `a', `an', `the', etc. which do not add much meaning to the text) as per the NLTK library
\item Lower casing Text: Conversion of the tweet text to lowercase, ensuring consistent, uniform handling of words
\item Removal of punctuations and twitter handles or usernames (`@')
\item Removal of non-alphanumeric characters excluding hashtags
\item Emoji Translation: Converting emojis to their corresponding textual representations
\item Contraction Expansion: Expansion of contractions to their full word forms, enhancing text clarity and comprehensibility.
\item Removal of URLs as they don't contribute to sentiment analysis
\end{itemize}
\subsection{Tokenization \& Lemmatization}
\textbf{Tokenization} refers to the process of breaking down of a sequence of text into smaller units or tokens. It is crucial for feature extraction, dimensionality reduction, normalization and semantic understanding. The tweets have been tokenized using the {\tt nltk.tokenize} package. \textbf{Lemmatization} is a text normalization technique, it is the process of reducing words to their meaningful base form (lemma) to ensure variants of a word are treated as a single item for analysis or retrieval purposes. We employed the {\tt WordNetLemmatizer} from the NLTK library for this task. We chose lemmatization over an alternative technique called stemming for a specific reason. Stemming involves chopping off the ends of words to reduce them to a common root. While this can be useful in some cases, it's not ideal when preserving the full meaning of words is important. Stemming can sometimes truncate words too aggressively, resulting in a loss of meaning. In our context, where understanding the context of text is crucial, lemmatization helps us retain more of the original word's meaning while still achieving normalization. Additionally, we integrated \textbf{Part-of-Speech (PoS) tagging} into our text processing pipeline which involves assigning a grammatical category (nouns, verbs, adjectives, etc.) to each word in the text. This further enhances our ability to understand the context and helps us make more precise decisions during lemmatization. 
\section{Models}
\subsection{LLM}
We have employed Large Language Models (LLMs) to address a particularly intricate and under-explored realm - multi-label classification, within the context of Social Media Tweets on vaccines. Our exploration involved the utilization of diverse prompting methods, such as chain of thought (CoT) style\textsuperscript{\cite{lu2022learn}}, zero-shot \& few-shot learning. Through rigorous experimentation, we discovered that an amalgamation of these prompting styles proved most effective for our objectives. Subsequently, we formulated a novel prompt template tailored specifically for multi-label text classification tasks with LLMs. We noted an enhancement in model performance when prompts requiring the model to generate explanations or reasonings were incorporated. The rationale behind the model's assignment of labels to a tweet exhibited a more coherent and meaningful pattern. A selection of instances of the model's reasoning is provided in Table 2 for reference.
\begin{table}[h]
\centering
\caption{Examples of LLM Outcomes}
%\begin{tabular}{|c|c|c|}
\begin{tabular}{|p{7cm}|p{3cm}|p{4cm}|}
\hline
\textbf{Tweet Text} & \textbf{Predicted Labels} & \textbf{Explanation Generated} \\
\hline
study link hpv vaccine to historically high infertility rate  \#spirituality \#self &  [`side-effect', \newline`religious']&\textit{The tweet expresses concern about the potential side effects of the HPV vaccine, specifically linking it to a historically high infertility rate. The tweet also mentions the keywords `spirituality' and `self', which suggests that the tweet is expressing opposition to the vaccine based on religious beliefs.}\\
\hline
am apreciat un videoclip pe  treatment and cure for gardasil hpv vaccine injury here leylas story &  [`side-effect']&\textit{The tweet mentions ``HPV vaccine injury" and ``Leyla\'s story" which suggests that the tweet is expressing worries about the adverse effects of the HPV vaccine. The keywords ``injury" and ``story" are associated with the ``Side-effect" label, which is why this label is applicable to the tweet.'}\\
\hline
oral polio vaccine infecting unvaccinated kid  via \#vet \#christian \#writer \#scifi \#trump \#maga &  [`ingredients', \newline`religious']&\textit{The tweet mentions the words `oral polio vaccine' and `unvaccinated kid', which suggests that the tweet is raising concerns about the ingredients present in the vaccine and its potential effects on unvaccinated children. The tweet also includes the words `Christian' and `Trump', which implies that the tweet is expressing opposition to the vaccine based on religious beliefs. The keywords `oral polio vaccine', `unvaccinated kid'}\\
\hline
\end{tabular}
\end{table}
\subsubsection{Template}
We used the \texttt{text-davinci-003} engine which belongs to the GPT 3.5\textsuperscript{\cite{gpt}} family, with configuration \texttt{temperature} = 0 and \texttt{max\_tokens} = 100. Our prompt template is outlined as follows:
\begin{lstlisting}[language=Python]
prompt = f"""
    Task: You are given a tweet on vaccines delimited by triple backticks(```). Your task is to pick the right Concern Labels for the tweet from the provided options. Also, explain the logical reasoning you took to choose a label, including the relevant words from the tweets. Learn from examples by understanding common words, keywords, and feelings associated with each label. Only choose labels you're confident about. Each label has a description and common keywords to help you understand the concern that the tweet is depicting.
    
    Concern List:
    1. Concern 1
    2. Concern 2
    ...

    ###
    Description of the 12 different vaccine concerns :
    1. Concern 1: Description of the Concern.
    ...

    ###
    Some of the keywords associated with tweets for each label:
    1. Concern 1: Keywords - present in a tweet for Concern 1.
    ...

    ###
    Examples:
    Tweet: "Training tweet "
    Concern: ['associated concerns']
    ...

    ###
    Format of response (Response should include Concern in the same format as in examples):
    Concern: [List of all the relevant applicable concern labels]
    Reasoning: [logical reasoning followed to decide each of the applicable labels]

    Tweet: ```{Tweet}```

    Note: Include only the most relevant concern labels in your response. Understand and analyze the sentiment and hidden meanings associated with the given tweet and compare it with the sentiments and keywords in the examples before responding. Comprehend the descriptions and keywords associated with each concern label and then assess the similarity with the given tweet's meaning and these concerns. Verify each and every label before responding to increase the prediction accuracy.
    """
\end{lstlisting}
\subsubsection{Proposed Prompt Template Approach}
In the context of addressing a 12-class multi-label classification problem, we propose a structured approach that includes the following essential components:
\newline
\textbf{Task Specification:} It is imperative to commence by precisely articulating the classification task under consideration. In our case, this involves the categorization of Social Media Tweets concerning vaccines into multiple concern categories.
\newline
\textbf{Option Presentation:} Subsequently, we put forth a comprehensive array of the 12 distinct concern labels for the model's consideration.
\newline
\textbf{Explanatory Context for Options:} To facilitate the model's comprehension, we provide detailed explanations and descriptions for each of the concern labels. These descriptions not only convey the essence of each concern but also elucidate common keywords (as per table 6\textsuperscript{\cite{caves2022}} along with some more additions to it), sentiments, or themes that are typically associated with them.
\newline
\textbf{Learning Through Examples:} An integral aspect of our approach involves training the model through illustrative examples. These examples consist of actual tweets paired with their corresponding concern labels. Additionally, we furnish comprehensive explanations accompanying these examples, elucidating the logical reasoning behind the assignment of specific labels to each tweet.
\newline
In the pursuit of optimizing the model performance, we maintain records of the specific adjustments and enhancements deemed necessary after testing. These modifications include various aspects, such as refining keyword lists, fine-tuning model parameters, and adapting the training strategy.
\newline
It is important to emphasize that our methodology transcends mere label assignment, we challenge the model to not only provide labels but also substantiate its choices through logical reasoning or explanations. This approach fosters a deeper understanding of the model's decision-making process, enabling us to elucidate the thought patterns underlying its multi-label classification predictions.
\subsection{Transformer based models}
\subsubsection{DistilBERT \& BERT}
Distillable Bidirectional Encoder Representations from Transformers is a widely used NLP model. DistilBERT base uncased\textsuperscript{\cite{distilbert}} (which has been used) is a lighter and faster variant of the BERT model that retains much of its capability to understand and generate human-like text.
\newline
One-hot encoding was applied on the pre-processed train set. For the first run, the 9921 tweets from the original training dataset were well-shuffled and split into training and validation sets in 80:20 ratio. The 7936 tweets and their ground truth labels were used to fine-tune the pre-trained model and the 1985 tweets (validation data) were used for evaluation purpose. Then, this model was run to predict the classes for the test set, with the hyper-parameters: {\tt max len} = 512, {\tt train batch size} = 16, {\tt learning rate} = 1e-05, {\tt num workers} = 2 and no. of {\tt epochs} = 10.
\newline
For the second run, the model was fine-tuned with the entire train set data (without any split). {\tt Dropout rate} of 0.5, {\tt weight decay} of 0.001, {\tt learning rate} of 1e-4, {\tt batch size} 16 and {\tt threshold value} of 0.5 were applied. Model was trained for 10 {\tt epochs}. These
hyper parameters values were carefully chosen to collectively contribute to improving the model training efficiency \& performance (by mitigating overfitting etc.).
\newline
BERT base uncased\textsuperscript{\cite{bert}} was also tried (on 80\% training \& 20\% validation data) with the hyper-parameters: {\tt max len} = 200, {\tt train batch size} = 8, {\tt learning rate} = 1e-05 and no. of {\tt epochs} = 10. BERT base uncased outperformed DistilBERT on the validation set, achieving a macro F1 score of 0.83 after 10 epochs of training. However, when applied to the test set, BERT faced challenges in predicting classes for a significant number of cases, unlike DistilBERT. The superior performance on the validation set can be attributed to BERT's larger capacity to learn from the training data. Yet, this advantage might've also made BERT more susceptible to overfitting, where it closely tailors its predictions to the training data, making it less adaptable to the new test data. It's possible that the hyper-parameters chosen during training favored BERT's performance on the validation set but were less suitable for the test set.
\begin{table}[h]
  \caption{Results on Validation Set (20\% of training data)}
  \label{tab:commands}
  \begin{tabular}{cccl}
    \toprule
    Model &Macro P & Macro R &Macro F1\\
    \midrule
    DistilBERT base uncased& 0.80& 0.70 &0.74\\
    \bottomrule
  \end{tabular}
\end{table}
\subsection{Traditional Methods}
We also explored some of the traditional machine learning techniques like Multinomial Naive Bayes (NB), Random Forest and Support Vector Machine (SVM) with TF-IDF vectorization on this task. However, it became evident during our experiments that these traditional methods struggled to effectively capture the complexity of the problem. The conclusion was drawn based on the validation scores, which consistently demonstrated limitations in their performance.
\begin{table}[h]
  \caption{Results on Validation Set (20\% of training data)}
  \label{tab:commands}
  \begin{tabular}{cccl}
    \toprule
    Model &Macro P & Macro R &Macro F1\\
    \midrule
    SVM with TF-IDF& 0.79& 0.34 &0.45\\
    MN-NB& 0.64& 0.31 &0.39\\
    RF with TF-IDF& 0.70& 0.18 &0.26\\    
    \bottomrule
  \end{tabular}
\end{table}
\subsubsection{Feature extraction}
\textbf{TF-IDF}: Term Frequency-Inverse Document Frequency is a widely used text vectorization technique in NLP and information retrieval. It helps in capturing the importance of words in a document (d) and across a corpus. TF measures the frequency of a term (t) within a document while IDF measures the importance of a term across a collection of documents.
\begin{equation}
TF(t, d) = \frac{\text{Number of times term } t \text{ appears in document } d}{\text{Total number of terms in document } d}
\end{equation}
\begin{equation}
IDF(t, D) = \ln\left(\frac{\text{Total number of documents in corpus } D}{\text{Number of documents containing term } t}\right)
\end{equation}
TF-IDF is the product of TF \& IDF which assigns a weight to each term in a document. Rare words that appear in specific documents will have a high TF-IDF score.
\begin{equation}
TF-IDF(t, d, D) = TF(t, d) \cdot IDF(t, D)
\end{equation}
\subsection{Classifier Chains}
Classifier chains\textsuperscript{\cite{cchains}}\textsuperscript{\cite{cchains2}} is a machine learning method for problem transformation in multi-label classification. It is an extension of Binary Relevance, where each label is treated as a separate binary classification problem.
\begin{table}[h]
  \caption{Results on Validation Set (20\% of training data)}
  \label{tab:commands}
  \begin{tabular}{cccl}
    \toprule
    Model &Macro P & Macro R &Macro F1\\
    \midrule
    Classifier Chains with TF-IDF& 0.74& 0.42 &0.48\\
    \bottomrule
  \end{tabular}
\end{table}
\section{Results}
\subsection{Metrics}
\textbf{Precision (P)}: It is the measure of the accuracy of positive predictions made by the model. It can be expressed as \[P = \frac{TP}{(TP + FP)}\] where TP and FP refer to True positives (number of correctly predicted positive instances) and False positives (number of incorrectly predicted positive instances) respectively.
\newline
\textbf{Recall(R)}: It answers what proportion of actual positives was identified correctly. It can be expressed as \[R = \frac{TP}{(TP + FN)}\] where FN refers to False negatives (number of incorrectly predicted negative instances).
\newline
\textbf{F-score(F1)}: It is the harmonic mean of precision and recall. It is a useful metric as it provides a more balanced summarization of model performance, considering both the P and R values. It can be expressed as \[F1 = 2 \times \frac{(P \times R)}{(P + R)}\]In terms of TP, FP and FN, the alternative equation can be \[F1 = \frac{TP}{(TP + \frac{1}{2} \times (FP + FN)}\]
These values are scaled between 0 and 1, with 1 signifying the highest achievable score and 0 denoting the poorest performance.
\newline
\textbf{Jaccard Score}: Also known as the Jaccard Index or Jaccard Similarity Coefficient, it is a measure of the similarity between two sets. It is defined as the size of the intersection of the sets divided by the size of the union of the sets. In context of our classification task, it can be represented as:
\[ J(y_{\text{true}}, y_{\text{pred}}) = \frac{|y_{\text{true}} \cap y_{\text{pred}}|}{|y_{\text{true}} \cup y_{\text{pred}}|} \]
where {\tt y\_true} and {\tt y\_pred} represent the ground truth \& predicted labels respectively. The Jaccard Score ranges between 0 and 1 where a score of 1 indicates that the predicted labels perfectly match the true labels (there are no false positives or false negatives) \& 0 means that there is no overlap between the predicted and true labels, indicating complete dissimilarity.
\subsection{Final Outcomes}
Out of all the methods we experimented with, the ones sent to the track include the final predictions made by the LLM (run 3) and DistilBERT (runs 1 \& 2, as mentioned in 5.2.1) on the test set.
\newline
The macro F1\textsuperscript{\cite{mf1}} and Jaccard\textsuperscript{\cite{jacc}} scores from the scikit-learn library are the metrics used to evaluate the performance of the model. The scores are computed based on the predicted labels on the test set. LLM achieved a slightly higher macro F1 score of 0.55, outperforming the transformer models like DistilBERT, albeit with a narrow margin.
\newline
Coming to the class-wise stats, we see that most of the models tend to struggle with two specific classes, `none' and `conspiracy'. Additionally, the `country' and `religious' classes, which have relatively fewer examples in the dataset, also tend to result in lower model performance.
\newline
We will be able to share the results for our other model predictions on the test data once we have the ground truth labels (which will be revealed after the conference concludes) for the test set.
\begin{table}[h]
  \caption{Final results on Test Set}
  \label{tab:commands}
  \begin{tabular}{cccl}
    \toprule
    Run File & Method &Macro F1 score & Jaccard score\\
    \midrule
    DataWarriors\_Run3.csv& LLM &0.55& 0.47\\
    DataWarriors\_Run2.csv& DistilBERT (no validation split) &0.54& 0.55\\
    DataWarriors\_Run1.csv& DistilBERT (with 20\% validation split) &0.53& 0.56\\
    \bottomrule
  \end{tabular}
\end{table}
\section{Limitations of the Study \& Future Aspects}
\begin{itemize}
\item Due to constraints of not having an OpenAI paid subscription, the model was trained using a subset of only 58 cases randomly chosen (thoughtfully selected in a way such that examples from all
12 classes were present) from the pre-processed train set. In the future, a more extensive exploration and understanding of diverse prompting strategies could yield even better results. \textit{It's important to highlight that LLM has proved it's effectiveness in terms of performance on this complex multi-label classification task. This suggests that with additional resources, there's a significant potential for further enhancing the model's performance.}
\item The exploration of advanced models, spanning a broader range of epochs and configurations, was hindered by limitations in GPU resources.
\item We observed that the model adhered to the provided concern labels. The structured prompts, incorporating keywords and notes, proved highly effective in reducing hallucinations. This approach empowered the model to develop a nuanced understanding of each concern label, and its responses were logically aligned with this comprehension. However, there are instances where the model exhibited hallucinations while generating explanations - false yet credible-sounding content. The fluency and quality of the generated non-factual content by the models are intriguing \& require further study to deduce patterns and understand the model's thought process. The findings from (Augenstein et al., 2023)\textsuperscript{\cite{augenstein2023factuality}} shed light on similar concerns related to large language model behaviour.
\item The incorporation of sentiments into responses, particularly within the domain of large language models, stands as an unexplored area. While it's a preliminary investigation, it provides insights into possible implications and suggests areas for future research.
\end{itemize}

\begin{acknowledgments}
We would like to extend our sincere gratitude to the organizers of the Artificial Intelligence on Social Media \href{https://sites.google.com/view/aisome/aisome}{(AISoMe)} track\textsuperscript{\cite{poddar2023aisome}} at the \href{http://fire.irsi.res.in/fire/2023/home}{FIRE 2023} conference, for hosting this brilliant \& insightful problem statement that served as a great learning experience. We're grateful for the opportunity to engage with it.
\end{acknowledgments}
\bibliography{sample-ceur}
\nocite{zhang2023stance}
\end{document}